\documentclass{article}

\usepackage{arxiv}
\usepackage[utf8]{inputenc} 
\usepackage[T1]{fontenc}    
\usepackage{hyperref}       
\usepackage{url}            
\usepackage{booktabs}       
\usepackage{amsfonts}       
\usepackage{nicefrac}       
\usepackage{microtype}      
\usepackage{lipsum}		
\usepackage{graphicx}
\usepackage{natbib}
\usepackage{doi}
\usepackage{amsmath}
\usepackage{xcolor}
\usepackage{hyperref}

\title{Explaining black boxes with a SMILE: Statistical Model-agnostic Interpretability with Local Explanations}

\date{October 11, 2023}	

\author{ 
    Koorosh Aslansefat \thanks{Corresponding Author}  \\
    School of Computer Science, University of Hull, UK \\
    \texttt{k.aslansefat@hull.ac.uk} \\
    \AND
    Mojgan Hashemian \\
    Independent Researcher, UK \\
	\AND
    Martin Walker \\
    School of Computer Science, University of Hull, UK \\
    \AND
    Mohammed Naveed Akram \\
    Fraunhofer IESE, Kaiserslautern, Germany \\
    \AND
    Ioannis Sorokos \\
    Fraunhofer IESE, Kaiserslautern, Germany \\
    \AND
    Yiannis Papadopoulos \\
    School of Computer Science, University of Hull, UK \\
}

\renewcommand{\shorttitle}{\textit{SMILE} Explainability}

\hypersetup{
pdftitle={A template for the arxiv style},
pdfsubject={q-bio.NC, q-bio.QM},
pdfauthor={David S.~Hippocampus, Elias D.~Striatum},
pdfkeywords={First keyword, Second keyword, More},
}

\begin{document}
\maketitle

\title{SMILE: Statistical Model-agnostic Interpretability with Local Explanations}


\begin{abstract} 
Machine learning is currently undergoing an explosion in capability, popularity, and sophistication. However, one of the major barriers to widespread acceptance of machine learning (ML) is trustworthiness: most ML models operate as black boxes, their inner workings opaque and mysterious, and it can be difficult to trust their conclusions without understanding how those conclusions are reached. Explainability is therefore a key aspect of improving trustworthiness: the ability to better understand, interpret, and anticipate the behaviour of ML models. To this end, we propose SMILE, a new method that builds on previous approaches by making use of statistical distance measures to improve explainability while remaining applicable to a wide range of input data domains.
\end{abstract}


\maketitle

\section{Introduction}
The global market for artificial intelligence (AI) has experienced significant growth in recent years, with the market expected to reach a value of over £150 billion by 2025. This rapid expansion has been driven by a range of factors, including the increasing availability of data, the development of more powerful hardware, and the growing demand for automation and decision-making support in a variety of industries \cite{kaynak2021golden}. A major focus in this area has been development of machine learning (ML) models, in which models like neural networks are trained on an initial, preprocessed set of data so that they can then perform that task on unseen data without guidance. ML is well-suited for applications in a wide range of domains, from healthcare (e.g. to support medical diagnosis), transportation and robotics (e.g. obstacle detection for autonomous vehicles or robots), and software development (e.g. as a developer copilot). 

A major obstacle to widespread adoption of ML, particularly in safety-critical areas, is the opaqueness of most ML models. They operate largely as black boxes, and once training is complete there is little control or insight into how they operate, which impacts confidence in their trustworthiness. For example, only a very brave doctor would recommend treatment solely on the basis of an ML medical diagnosis when there are no clues as to how that diagnosis was reached. Indeed, in one famous example, an ML system to detect malignant skin cancers would often incorrectly interpret images with rulers as malignant, since positive examples in the training dataset frequently featured a ruler for measurement \cite{ruler_error}. 

To address this, there is increasing focus on explainability and interpretability. These terms refer to the ability of an AI system to provide clear and understandable explanations for its decisions. The aim is to identify the degree to which features of the input are indicative or counter-indicative of a given decision. Such information is useful both to give developers greater insight when testing the model and also to give end users more confidence in its predictions. Explainability is particularly important in helping to detect discriminatory biases (whether intentional or not) that may otherwise go unnoticed in black box ML models \cite{adadi2018peeking}.

SMILE (Statistical Model-agnostic Interpretability with Local Explanations), the technique we propose in this article, builds upon existing explainability techniques by making use of statistical distance measures. \textcolor{black}{The goal is to improve the consistency and accuracy of the explanations by better representing the impact of different features on the decision and reducing sensitivity to sharp discontinuities in the input that might otherwise yield misleading explanations. SMILE is applicable across a range of input data types, like tabular data and images.}

\section{Explaining Explainability}

The most common categorization of explainability methods distinguishes between those methods which are model-specific, i.e., which depend on knowledge of a particular model's characteristics and/or its development process, and those which are model-agnostic, i.e., require no internal knowledge of the black box model and can be applied post-hoc.

Model-specific approaches, such as the Bayesian List machine \cite{letham2012building}, leverage their understanding of the model to yield a deeper insight into its behaviour. The trade-off is limited applicability, since they can only be applied to those models they are designed for. By contrast, model-agnostic techniques do not require prior knowledge and can be applied to a wide range of ML models. Most model-agnostic approaches are post-hoc in nature (e.g. \cite{lime}, \cite{NIPS2017_7062}, \cite{zhou2021s}, \cite{zhao2021baylime}), being applied after the model has been trained by correlating the impact of features on the output for a given input. Most model-specific approaches are ad-hoc techniques that incorporate explainability into the model itself (such as \cite{caruana2015intelligible} and
\cite{letham2012building}). However, post-hoc model-specific approaches (like \cite{sundararajan2017axiomatic} and \cite{rieger2020interpretations}) also exist.

A more comprehensive classification and review of explainability techniques can be found in \cite{gunning2019darpa}, while the authors of \cite{arrieta2020explainable} provide a detailed overview of the existing techniques available for explainable AI. 

For numerous reasons, including the wider applicability and the desire to keep ML technology proprietary, model-agnostic approaches have proven to be the most common. Of these, two of the most widespread such methods are LIME and SHAP. LIME (Local Interpretable Model-agnostic Explanations) \cite{lime} is intended to explain the decision of a black box classifier by fitting a local, interpretable model around the prediction. Essentially, it uses random perturbations around the input data to collect information about how different features contribute to the output decision, weighted by distance, and in doing so attempts to mimic the decision-making of the black box ML model. 

However, because of the randomness inherent in the perturbations, LIME has been shown to generate inconsistent results over multiple explanations, giving rise to techniques such as BayLIME \cite{zhao2021baylime} and S-LIME \cite{zhou2021s} that attempt to address this issue. BayLIME uses Bayesian-principled prior knowledge to generate explanations that are more consistent over multiple iterations, \textcolor{black}{while S-LIME leverages the central limit theorem to determine the necessary number of perturbation points for stable explanations.} 

SHAP (SHapley Additive exPlanations) \cite{NIPS2017_7062} is also intended to overcome the limitations found in LIME. It attempts to provide a more unified approach that takes advantage of concepts from game theory to better match human intuition in its explanations. SHAP achieves this by combining LIME's linear weighting model with Shapley values, which are used as part of a game theory model that treats each feature as a player attempting to maximise their gain (impact) on the outcome.

\section{Time to SMILE}
SMILE attempts to address the problems of inconsistency and sensitivity to local fluctuations around the input by making use of more sophisticated statistical methods. Rather than comparing a single perturbed value and the input sample, as most other approaches do, SMILE instead compares localised regions against each other. This yields a clearer explanation of how regional features impact the prediction -- like viewing the input space through a magnifying glass rather than a microscope.

SMILE is derived from LIME and thus the general approach is similar. The difference lies in the nature of the measures used to evaluate the contribution of different features. To explain how it works, consider a simple 2D tabular dataset for a black-box ML algorithm intended to classify data into one of two sets based on two features represented by two axes, x1 and x2.

\begin{figure*}
\centerline{\includegraphics[width=35pc]{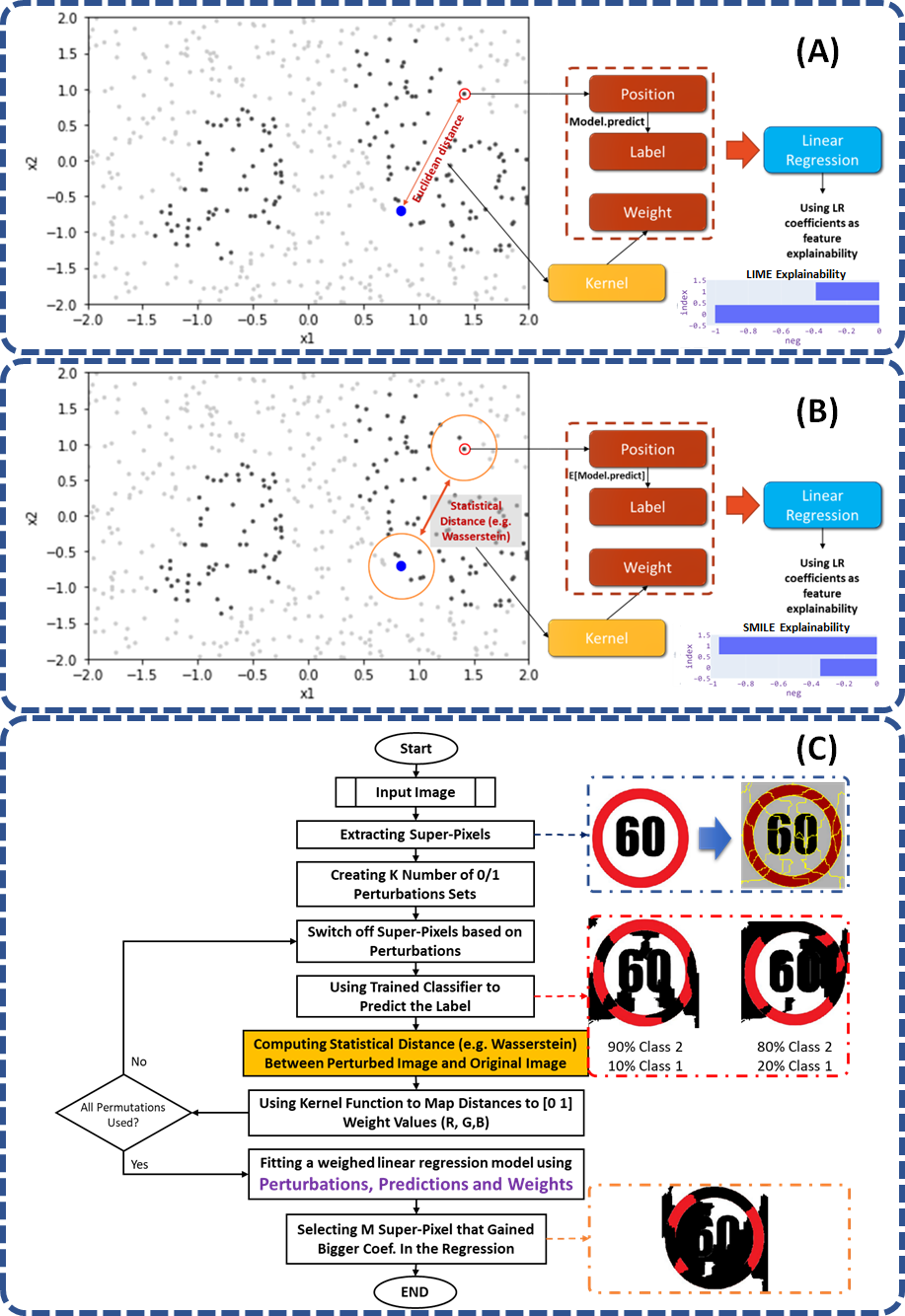}}
\caption{\textcolor{black}{(A) An example of how LIME works, (B) An example of how SMILE works, (C) SMILE flowchart for explaining image-based classification or regression}}
\label{fig1}
\end{figure*}

To explain how the model arrives at its decisions, both LIME and SMILE first generate randomly perturbed inputs around the given input sample and feed them to the black box classifier to obtain the predicted labels. LIME then calculates the Euclidean distance or cosine distance (depending on the type of data) between the sample and each randomly generated input. Using a proper kernel function, these calculated distances can be mapped to weights for each feature. Finally, using the set of generated data, predictions, and weights from the previous steps, a weighted linear regression model is trained. The coefficients of this model can then be used as the local explanation for each feature and each class. Figure \ref{fig1}-(A) illustrates the overall view of this procedure.

SMILE works similarly, but instead of calculating Euclidean distance between the sample and perturbed data, Empirical Cumulative Distribution Function-based (ECDF) statistical measures are used. This entails generating further random samples around both the original input sample as well as each perturbed input and then calculating the statistical distance between the two sets of samples. The overall SMILE process (illustrated in Figure \ref{fig1}-(B)) is therefore:

i) Generation of randomly perturbed data in the local region of the input sample in two stages. 

Stage 1 (Primary Perturbation): For each instance \( x \), generate \( N \) perturbed samples by introducing variations using a Gaussian distribution: 
\begin{equation}
\label{eq1}
    x_i^{(1)} \sim \mathcal{N}(x, \sigma_1^2) \quad \text{for} \quad i = 1, \ldots, N
\end{equation}
where \( x \) is the original instance and \( \sigma_1^2 \) is the variance of the Gaussian distribution used for primary perturbation.
Stage 2 (Local Perturbation): For each primary perturbed sample \( x_i^{(1)} \), generate \( M \) secondary perturbed samples by using \( x_i^{(1)} \) as the center of a Gaussian distribution:
\begin{equation}
\label{eq2}
 x_{ij}^{(2)} \sim \mathcal{N}(x_i^{(1)}, \sigma_2^2) \quad \text{for} \quad j = 1, \ldots, M 
 \end{equation}
 where \( x_i^{(1)} \) is the primary perturbed sample and \( \sigma_2^2 \) is the variance of the Gaussian distribution used for secondary perturbation, typically smaller than \( \sigma_1^2 \) to ensure small variations around \( x_i^{(1)} \).
Thus, for each original instance \( x \), there will be a total of \( N \times M \) secondary perturbed samples.

ii) \textcolor{black}{Calculation of the expected value of predictions by feeding a set of Local Perturbations to the black box ML model.}

iii) Calculation of the statistical distances between the sets from (i) and (ii). For example, equation \ref{eq3} demonstrates the Wasserstein Distance (WD) for two univariate distributions of $F_1\ (x)$ and $F_2\ (x)$ where $F_1\ (x)$ is the ECDF of the target point and the local distributed samples around it, and $F_2\ (x)$ is the ECDF of the set of clustered random points with respect to their class labels. For multi-variate datasets, the WD can be calculated for each feature.

\begin{equation}
\label{eq3}
    WD=\ \int_{-\infty}^{+\infty}\left|F_1\left(x\right)-F_2\left(x\right)\right|dx
\end{equation}

\textcolor{black}{
v) Using an exponential kernel function, the statistical distances are mapped to weights:
\begin{equation}
\label{eq4}
w_i = \exp(-\frac{WD^2}{\sigma^2})
\end{equation}
}
iv) Finally, as in LIME, the features, predicted labels, and weights from the previous steps are used to train a weighted linear regression model. The coefficients of this model help explain the contribution of each feature to each class.

In order to ensure both interpretability and local fidelity, we must minimize \( L(f, g, \pi_x) \) while having \( \Omega(g) \) be low enough to be interpretable by humans. The explanation produced by SMILE is obtained by solving the following optimization problem:
\begin{equation}
\label{eq5}
 \min_{g \in \mathcal{G}} L(f, g, \pi_x) + \lambda \Omega(g) 
\end{equation}

 where \( g \) represents the interpretable model from a class of models \( \mathcal{G} \), such as linear models or decision trees. The model \( f \) is the complex, black-box model we aim to explain. The term \( L(f, g, \pi_x) \) measures the fidelity of \( g \) in approximating \( f \) within a locality defined by \( \pi_x \), a proximity measure around the instance \( x \). The complexity of the explanation \( g \) is quantified by \( \Omega(g) \), where a simpler \( g \) will have a lower value. The parameter \( \lambda \) serves as a regularization term, balancing the trade-off between fidelity (captured by \( L \)) and interpretability (captured by \( \Omega \)).

In the above-mentioned procedure and most of the examples, we have used Wasserstein Distance because it calculates the area between two ECDFs and thus captures the geometry-related features of the distributions. However, a range of other ECDF-based measures are also used in SMILE, including Kolmogorov-Smirnov (maximum distance between two ECDFs), Kuiper (combines maximum and minimum distance), and Cramér-von Mises (which accumulates the distance at regular intervals). ECDF-based measures are used because they are more sensitive to the geometry of the distribution in comparison to PDF-based statistical distances like the Bhattacharyya distance, which mostly rely on mean and variance values.

SMILE is not limited to 2D tabular data and can also be applied to image or text-based datasets. For these types of data, LIME must generate vectors representing original and perturbed images/text and calculate the cosine distance between them. By contrast, SMILE can continue to use ECDF-based statistical distances, which our experiments so far have shown produces more accurate explanations. 

As an example, we have applied SMILE to the problem of identifying traffic signs (e.g. for self-driving vehicles or driver assistance). This process is illustrated in Figure \ref{fig1}-(C). The goal is to explain the decisions of a pre-trained black box image classifier by indicating which parts of the input image led it to recognising a given traffic sign. For each image, super-pixels (groups of pixels with common characteristics) are extracted and used to reduce the number of calculations required and increase performance.

Based on the number of detected super-pixels, K binomially-distributed random perturbation vectors are generated. Each element of a perturbation vector represents the status of a corresponding super-pixel (0: means the super-pixel should be excluded, and 1: means the super-pixel should be included). K perturbed images can thus be generated with different combinations of super-pixels and predictions obtained for each one (i.e., the predicted traffic sign). 

ECDF-based statistical distance measures are then used to calculate the distance of each perturbed image from the original sample, similar to the approach used in \cite{aslansefat2021toward}. As before, this then enables us to train a weighted linear regression model as a surrogate model. By sorting the coefficients of this surrogate model, the M super-pixels with the highest coefficients can be selected and presented as the explanation, showing which parts of the image have the greatest impact on the ML model's decision.

The use of ECDF-based calculations rather than Euclidean (or cosine) measures increases the computational complexity over LIME, but by sampling more data points, SMILE captures a more accurate picture of how each feature impacts the ML model's decision, especially when the data point is on a decision boundary. 

\section{Something to SMILE about}
We have performed a number of experiments to evaluate the performance of SMILE compared to common alternatives (LIME and SHAP and, for images, BayLIME). 
\subsection{\textcolor{black}{Human Intuition}}
One way to evaluate an explanation is to compare the outcome to human intuition: a trustworthy explanation should be consistent with the explanations of humans who have proper expert knowledge and an understanding of the model. For example, if both an ML model and a doctor assess the same MRI scan image, both should highlight the same features of the image as the reason for their diagnosis.

\begin{figure*}
\centerline{\includegraphics[width=37pc]{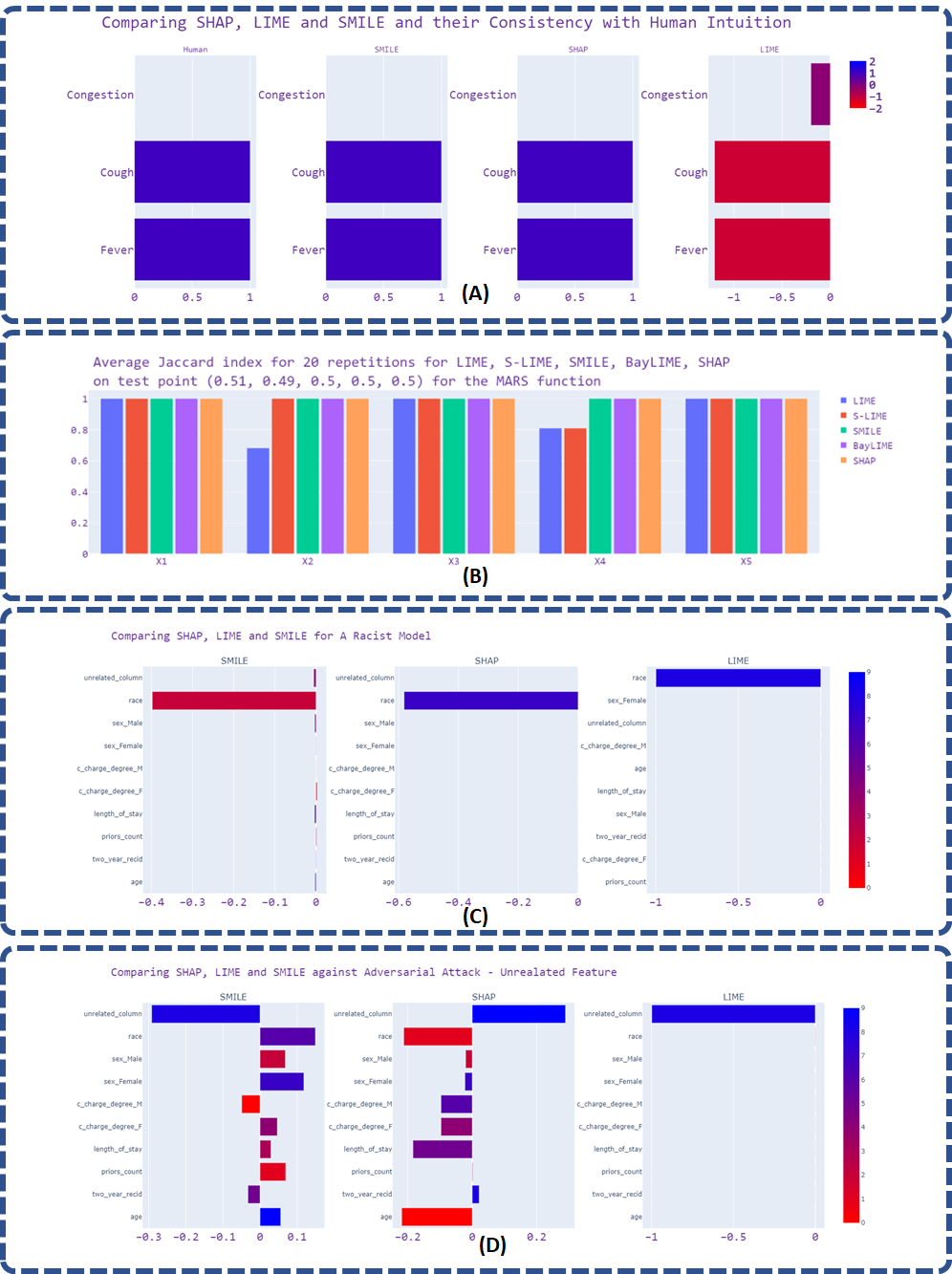}}
\caption{\textcolor{black}{Comparing SHAP, LIME and SMILE (A) and their Consistency with Human Intuition (Adapted from Lundberg \& Lee (2017) and compared with SMILE), (B) COMPAS Dataset and XgBoost Model without Adversarial Attack, (C) against Adversarial Attack and a Racist Model, (D) against Adversarial Attack and a Model with Unrelated Feature [X-AXIS are Explainability Values]}}
\label{fig2}
\end{figure*}

In order to test this, we have reproduced the experiment in \cite{NIPS2017_7062}. Figure \ref{fig2}-(A) shows the comparison of human feature impact estimates versus those provided by LIME, SHAP, and SMILE. The experiment is based on a model that determines a sickness based on two indicative symptoms, fever and cough, with a third unrelated symptom, congestion, also included. As can be seen in the figure, both SMILE and SHAP are consistent with the human explanations, providing positive values for fever and cough and zero for congestion. However, LIME's explanations were not consistent: LIME not only provided negative values for fever and cough but also assigned small negative values for congestion, which is an unrelated symptom. 

\subsection{\textcolor{black}{Quantifying Explanation Stability}}

\textcolor{black}{To assess the stability of the generated explanations, we employ the Jaccard index, a statistical measure designed to evaluate the similarity and diversity of sample sets. Given two sets \( A \) and \( B \) (in our context, these sets represent selected features from LIME), the Jaccard coefficient, \( J \), is defined as:}
\textcolor{black}{
\begin{equation}
\label{eq5_1}
J(A, B) = \frac{|A \cap B|}{|A \cup B|} 
\end{equation}
}
\textcolor{black}{
where \( |A \cap B| \) is the size of the intersection of sets \( A \) and \( B \) and \( |A \cup B| \) is the size of the union of sets \( A \) and \( B \). This coefficient provides a value between 0 (indicating no overlap) and 1 (indicating complete overlap), offering a quantitative measure of the stability and consistency of the explanations generated.}

\textcolor{black}{In our analysis, we employ a modified version of the function presented in \cite{zhou2021s}, originally used to test the MARS algorithm. This function serves as our black box model, allowing us to ascertain the true local weights of variables. The function is defined as: }
\textcolor{black}{
\begin{equation}
\label{eq5_2}
 f(\mathbf{x}) = 10 \sin(\pi x_1 x_2) + 20(x_3 - 0.05)^2 + 5x_4 + 5x_5 
 \end{equation}
 }
 \textcolor{black}{
 where \( \mathbf{X} \sim \mathcal{U}([0, 1]^5) \). The chosen test point \( \mathbf{x} \) is (0.51, 0.49, 0.5, 0.5, 0.5). Figure \ref{fig2}-B displays average Jaccard indices for 20 repetitions of LIME, S-LIME, SMILE, BayLIME, SHAP on test point (0.51, 0.49, 0.5, 0.5, 0.5) for the MARS function. The results suggest LIME and S-LIME experience some instability while the others provided stable results.} 

\subsection{\textcolor{black}{Explainability of a Biased Model}}
 
\textcolor{black}{To ascertain the ground truth, one approach is to intentionally bias a model towards a specific feature and then assess how various explainability methods react. For instance, Figure \ref{fig2}-(C) presents a model with racial bias. Notably, all three explainers identify race as the predominant feature. Within SMILE, other features register minimal values, attributable to the influence of local perturbations, as depicted by the two circles in Figure \ref{fig1}-(B). Practically speaking, there exists an inherent trade-off: on one hand, this provides robustness against manipulative tactics like adversarial attacks, and on the other, it can be less precise in pinpointing specific feature explanations. This balance can be adjusted by varying the number of local perturbations within the algorithm.}

\subsection{\textcolor{black}{Adversarial Attack on Explainability}}

As discussed in \cite{NIPS2017_7062}, models can be designed to fool explainability methods into highlighting an unrelated feature using adversarial attacks. The same approach is used to test SMILE as well. As illustrated in Figure \ref{fig2}-(D), LIME has been fooled by the attack while SHAP and SMILE both show a degree of robustness to the attack. 

\textcolor{black}{To quantify this robustness, one can calculate the ratio of the value from the unrelated column to the total values provided by the explainer X as: ${\left| X_{\text{index\_unrelated}} \right|}$ $/$ ${\sum_{i=1}^{n} \left| X_i \right|}$. A result of zero indicates full robustness against the attack, while a score of one implies the explainer was entirely deceived. Based on this metric, the outcomes for SMILE, SHAP, and LIME are 0.321, 0.249, and 0.996 respectively, indicating that SHAP is the most resilient here, with SMILE following closely behind.}

\subsection{\textcolor{black}{SMILE for Image Explainability}}
Explainability in images is particularly challenging compared to simple tabular data because the features are all part of the image rather than numbers conveniently separated into discrete sets. Explaining the diagnosis of an illness from a scan image means breaking down the image into those parts that either support or contradict the given diagnosis, whereas the task of explaining a diagnosis made on the basis of variables representing discrete symptoms is much smaller in scope. 

\textcolor{black}{The current iteration of LIME employs a perturbation function that toggles super-pixels on and off, subsequently comparing the resultant perturbation vectors via cosine metrics. In contrast, SMILE's approach of comparing original and perturbed images not only enhances accuracy but also offers flexibility in the choice of perturbation methods, such as the addition of noise. This distinction suggests that statistical distance measures, as utilized in SMILE, can potentially convey richer information than cosine distance alone.}

\begin{figure*}
\centerline{\includegraphics[width=35pc]{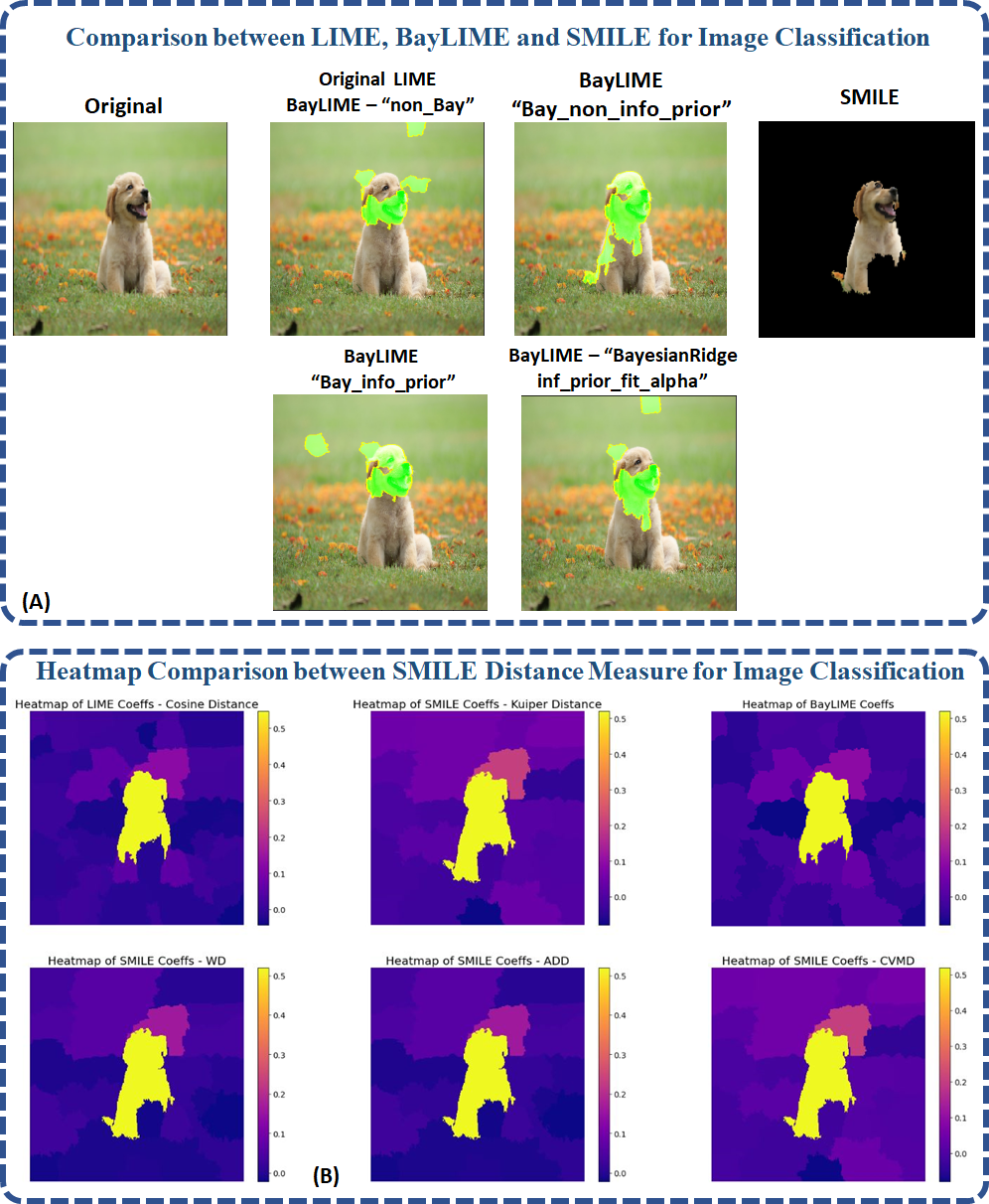}}
\caption{\textcolor{black}{ (A) Comparing SMILE with LIME and BayLIME for dog classification and its explainability for inceptionV3 model. (B) Heatmap Comparison between 1) LIME, 2) SMILE with Kuiper Distance, 3) BayLIME, 4) SMILE with Wasserstein Distance, 5) SMILE with Anderson-Darling Distance (ADD), 6) SMILE with Cramér-von Mises Distance (CVMD) -- for Image Classification.}}
\label{fig3}
\end{figure*}

As an illustrative example, Figure \ref{fig3} presents the results of a dog classification using the Inception\_V3 model, as described in \cite{zhao2021baylime}. Figure \ref{fig3}-(A) delineates a comparison between LIME, BayLIME and SMILE. BayLIME is another extension of LIME that focuses on improving the surrogate model by replacing its weighted linear regressor model with a Bayesian regressor model. The top-left image is the original given to the model. The next four images show four different BayLIME settings (from \cite{zhao2021baylime}) -- non\_Bay, which is similar to the original LIME approach; Bay\_non\_info\_prior, Bay\_info\_prior, and BayesianRidge\_inf\_prior\_fit\_alpha. On the right is the SMILE result. As can be seen, SMILE largely outperforms both the original LIME and BayLIME, successfully identifying the dog as the most important feature while not identifying any irrelevant sections of the image as being important.

Figure \ref{fig3}-(B) shows the heat-map for SMILE explainability with different measures: cosine, Kuiper, Kolmogrov-Smirnov, Wasserstein, Anderson Darling and \textcolor{black}{Cramér-von Mises}. \textcolor{black}{The provided heat-map shows that ECDF-based explainability can provide more accurate results: compared to LIME or BayLIME, all of the different measures in Figure \ref{fig3}-(B) capture the bulk of the dog's shape with minimal unrelated background, correctly indicating the most important elements for classification of the dog. For a quantitative comparison, two parameters of coverage and weighted coverage have been defined as follows.} 

\textcolor{black}{
To quantify the coverage, a simple coverage estimation is used. 
 \begin{equation}
\label{eq6}
\begin{aligned}
coverage = \frac{1}{\sum_{x=0, y=0}^{w,h} M_{x,y}=coi} \times \\
\left( \sum_{x=1, y=1}^{w, h}[X_{x,y} > 0 \cap M_{x,y} = coi] \right. \\
\left. - \sum_{x=1,y=1}^{w,h}[X_{x,y} > 0 \cap M_{x,y}\neq coi] \right)
\end{aligned}
\end{equation}
}

\textcolor{black}{
where x is the number of pixels in the X-direction for an image of width of w, y is the number of pixels in the Y-direction for an image of height of h, X is the explanation, M is ground truth, and coi is the class of interest.
}

\textcolor{black}{
To take weight into account, we quantify the coverage with a weighted coverage estimation. We first modify the label as follows:
\begin{equation}
\label{eq7}
 M_{x,y} = \begin{cases}
1 & \text{ if } M_{x,y} = coi\\ 
-1 & \text{ if } M_{x,y} \neq  coi 
\end{cases}  
\end{equation}
}
which can be used to obtain the weighted coverage as: 
\begin{equation}
\label{eq8}
coverage_w = \frac{\sum_{x=1, y=1}^{w, h}[X_{x,y} * M_{x,y} ]}{w*h} 
\end{equation}

For SMILE, the coverage associated with the true label stands at 0.58255, while its weighted coverage is 0.02818. In contrast, BayLIME with no prior info registers a coverage of 0.50795 and a weighted coverage of 0.026784. LIME's metrics are 0.50795 for coverage and 0.028852 for weighted coverage. These figures suggest that SMILE exhibits a marginally superior performance when compared to both LIME and BayLIME for this specific example.

\subsection{Capabilities}
In the numerical result section, the capabilities of SMILE has been measured. In this section, those capabilities can be summarised as follows:
\begin{itemize}
	\item The proposed SMILE algorithm utilizing LIME procedure to use statistical distance measures and provide new type of explanation.
	\item This algorithm is completely model-agnostic, and it can work as an explainer for different ML/DL algorithms (for both classification and regression tasks).  
	\item In SMILE, we almost have the capabilities of LIME, and also we can utilize SHAP plotting capabilities to use them for SMILE. In addition, SMILE uses statistical properties that provide more robust explanations. 
	\item The SMILE explanation is compared with LIME and SHAP for Boston dataset and in for a hypothetical flu dataset the SMILE results were in-line with human feature impact estimates.
	\item The SMILE robustness against adversarial attacks has tested through two different examples. Although it was not 100 percent robust, but it performs better than LIME and SHAP.
\end{itemize}

\section{Conclusion} \label{section-conclusion}
Given the inexorable rise of AI and machine learning in particular, the issue of trustworthiness will only continue to become more critical. Explainability is one of the key tools in addressing this deficit in trustworthiness, and to ensure that it does not lag behind, continual development is vital to both expand the range of tasks that can be explained and to ensure that those explanations are consistent, accurate, and robust.

SMILE is intended as a new step along this path of development. It is model-agnostic and applicable to tabular, image, and text data, and we have evaluated its capabilities and performance using a variety of examples and test datasets, \textcolor{black}{though we recognize the need for additional experiments to further measure performance and generalizability. However, the experiments so far show that SMILE offers modest but useful advantages over comparable techniques such as LIME. We are also working on extending SMILE to apply to more complex forms of data, such as graph-based datasets and time-series data (such as the readouts of medical monitoring equipment). Given these considerations, our future work will be dedicated to a thorough evaluation of SMILE, testing its robustness and generalizability in various scenarios.}

This enhanced capability does come at a cost, however, and SMILE has a higher computational complexity compared to LIME or SHAP due to the use of statistical distance measures. For example, if we compare the distance measures used in SMILE and LIME for images, one can argue that the computation complexity of cosine distance is $O(n)$ while the computation complexity of Wasserstein is $O(n^{3} \log n)$. This additional complexity is noticeable but not prohibitive; for example, the dog image above may take around a minute to analyse for LIME and 4-5 minutes for SMILE. We are investigating ways to optimise the process, e.g. by offloading some of the calculation to GPUs or by making use of faster alternative measures like Sinkhorn, which is a fast approximation of Wasserstein. 

One other interesting issue we intend to explore further is that of adversarial attacks: attempts to fool explainability techniques into thinking a deliberately biased model is in fact innocuous \cite{slack2020fooling}. For example, a model could base its decisions on protected characteristics like race or gender but appear to the explainability technique as if it is making decisions on other features. An explanation can only improve the trustworthiness of the model if it is itself trustworthy, and so this is an important issue. SMILE's ability to assess a greater portion of the input data potentially offers a way to improve the robustness of explanations, though further experiments are required before any firm conclusions can be drawn.

\section*{CODE AVAILABILITY}
Regarding the research reproducibility, codes and functions supporting this paper are published online at GitHub: \href{https://github.com/Dependable-Intelligent-Systems-Lab/xwhy}{https://github.com/Dependable-Intelligent-Systems-Lab/xwhy}.

\section*{ACKNOWLEDGMENTS} \label{section-acknowledgements}
This work was supported by the Secure and Safe Multi-Robot Systems (SESAME) H2020 Project under Grant Agreement 101017258.


\bibliographystyle{unsrtnat}

\end{document}